\documentclass[conference]{IEEEtran}
\IEEEoverridecommandlockouts
\usepackage{cite}
\usepackage{amsmath,amssymb,amsfonts}
\usepackage{algorithmic}
\usepackage{graphicx}
\usepackage{textcomp}
\usepackage{xcolor}
\usepackage{float}
\usepackage{booktabs}
\usepackage{caption}
\usepackage{subfigure}
\usepackage{amsmath}
\usepackage{epstopdf}
\usepackage{epsfig}
\usepackage[ruled]{algorithm2e}
\usepackage{multirow}
\usepackage{hyperref}
\usepackage{xcolor}
\usepackage{colortbl} 
\usepackage{hhline} 
\usepackage{longtable}
\usepackage{lscape}
\usepackage{float}

\hypersetup{hidelinks}

\def\BibTeX{{\rm B\kern-.05em{\sc i\kern-.025em b}\kern-.08em
		T\kern-.1667em\lower.7ex\hbox{E}\kern-.125emX}}
\begin{document}
	
	\title{Multimodal Large Language Models-Enabled UAV Swarm: Towards Efficient and Intelligent Autonomous Aerial Systems}
	
	\author{Yuqi Ping, Tianhao Liang, Huahao Ding, Guangyu Lei, Junwei Wu, Xuan Zou, Kuan Shi, Rui Shao,\\ Chiya Zhang,  Weizheng Zhang, Weijie Yuan and Tingting Zhang
		\thanks{Yuqi Ping, Tianhao Liang, Huahao Ding, Guangyu Lei, Junwei Wu, Rui Shao, Chiya Zhang,  Weizheng Zhang and Tingting Zhang are with the College of Informatics, Harbin Institute of Technology, Shenzhen 518000, China (e-mail: pingyq@stu.hit.edu.cn, liangth@hit.edu.cn, 190210108@stu.hit.edu.cn, GuangyuLei@stu.hit.edu.cn, 220210419@stu.hit.edu.cn, shaorui@hit.edu.cn,  zhangchiya@hit.edu.cn, zhangweizheng@hit.edu.cn, zhangtt@hit.edu.cn); Xuan Zou and Kuan Shi are with the Key Laboratory of Forest and Grassland Fire Risk Prevention, Ministry of Emergency Management, China Fire and Rescue Institute, Beijing 102202, China (e-mail: icy66@126.com, 149991604@qq.com);  Weijie Yuan is with Southern University of Science and Technology, Shenzhen 518000, China (e-mail: yuanwj@sustech.edu.cn).
		}
	}
	
	\maketitle
	
	\begin{abstract}
		
	Recent breakthroughs in multimodal large language models (MLLMs) have endowed AI systems with unified perception, reasoning and natural-language interaction across text, image and video streams. Meanwhile, Unmanned Aerial Vehicle (UAV) swarms are increasingly deployed in dynamic, safety-critical missions that demand rapid situational understanding and autonomous adaptation. This paper explores potential solutions for integrating MLLMs with UAV swarms to enhance the intelligence and adaptability across diverse tasks. Specifically, we first outline the fundamental architectures and functions of UAVs and MLLMs. Then, we analyze how MLLMs can enhance the UAV system performance in terms of target detection, autonomous navigation, and multi-agent coordination, while exploring solutions for integrating MLLMs into UAV systems. Next, we propose a practical case study focused on the forest fire fighting. To fully reveal the capabilities of the proposed framework, human-machine interaction, swarm task planning, fire assessment, and task execution are investigated. Finally, we discuss the challenges and future research directions for the MLLMs-enabled UAV swarm. An experiment illustration video could be found online at https://youtu.be/zwnB9ZSa5A4.
	\end{abstract}
	

	\section{Introduction}
	Unmanned Aerial Vehicles (UAVs), with their compact size, low cost, and ease of deployment, have promoted the revolutionary advancements across various fields, attracting considerable attention from academia and industry. For example, UAVs can enable rapid and accurate package delivery for the last-mile logistics and extend the network coverage in disaster-stricken or remote areas. They can also aid in infrastructure inspections by detecting safety hazards in power lines, railways, and bridges, and support disaster rescue operations with quick access and real-time data for search and rescue efforts.
	
	However, current UAV applications still encounter many challenges. Most UAVs rely on the manual control, leading to high labor costs and potential safety risks due to limited environmental awareness and decision-making capabilities  of operators\cite{uav_survey}. Additionally, manual control requires real time transmission of high-precision, low-latency perception and control data, placing significant demands on the bandwidth and reliability of the UAV communication links. Existing UAV systems often struggle to balance the conflicting requirements of low power consumption and high communication performance, particularly in wide-area remote control applications. As a result, the autonomous control of UAVs has emerged as a key research focus.
	
	Several studies have investigated the integration of sensors and controllers on UAVs for autonomous navigation, enabling tasks such as cave exploration and disaster assessment \cite{uav_survey}. However, these approaches often rely on pre-programmed task flows and rule-based libraries, restricting them to executing predefined actions. This makes it difficult for them to adapt to diverse real-world task demands or respond effectively to unexpected situations. Furthermore, the task construction demands that operators have extensive expertise and requires considerable development time, making human-machine interaction complicated. Additionally, conventional perception and detection algorithms are typically constrained by predefined environments and limited training data, which hinder their adaptability and generality. These drawbacks limit the applications of autonomous UAV systems in real-world scenarios. Consequently, integrating multimodal large language models (MLLMs) into the UAV offers a promising way to surmount these challenges.
	
	With the rapid development of AI technologies, especially the emergence of MLLMs like GPT-4o and Qwen-VL, AI systems have reached the new levels of capability. These models possess strong natural language reasoning abilities and cant efficiently integrate data from multiple modalities, enabling them to understand and process information from various domains and perspectives deeply \cite{MLLMreview}. Consequently, integrating MLLMs into UAVs provides a new paradigm for human-machine interaction and helps UAVs efficiently integrate data from multiple sensors. By leveraging technique such as retrieval-augmented generation (RAG) and chain-of-thought (CoT), MLLMs-enabled UAV system can make decisions based on existing knowledge and experience, overcoming the limitations of task flows and enhancing the robustness and generalization of UAVs in complex scenarios.
	
	This paper explores how MLLMs can enhance the performance and scalability of UAV systems. The organization of this article is listed as follows. We first outline the fundamental structures and functions of UAVs and MLLMs. Next, we examine the potential of MLLMs in enhancing the task execution capabilities of UAV systems by reviewing current related works. Subsequently, we present a practical  scenario in forest firefighting, demonstrating the real-world application of MLLMs-enabled UAV systems in task execution, swarm collaboration, and system deployment. Finally, we discuss the challenges and future research directions and draw conclusions.
	\begin{figure}[htbp]
		\centering
		\includegraphics[width=0.8\columnwidth]{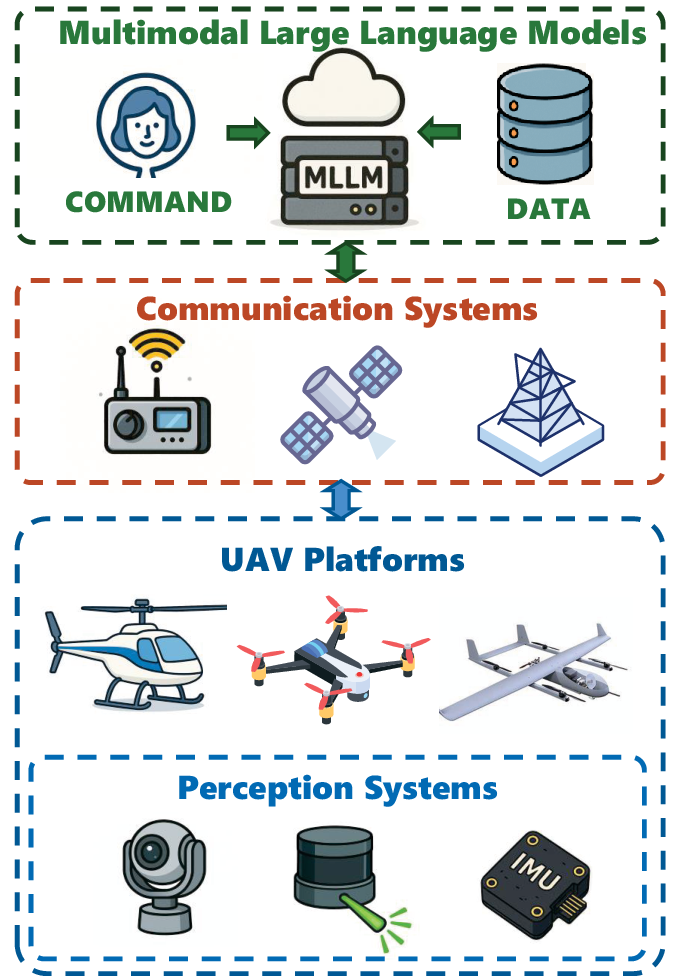}
		\caption{The general MLLMs-enabled UAV system architecture.}
		\label{system}
	\end{figure}
	
	\section{Overview of UAV Systems and MLLMs}
	The general MLLMs-enabled UAV system architecture is illustrated in Fig. \ref{system}, comprising four main modules: the MLLMs, the communication systems, the UAV platforms, and the perception systems. The communication systems ensure reliable information transmission between the MLLMs and the UAV. The UAV platforms execute task operations, while the perception system is responsible for real-time sensing and environmental feedback. A detailed introduction of each component will be provided in the following.
	\subsection{UAV Platforms}
	UAV platforms can be broadly categorized based on aerodynamic characteristics, propulsion methods, and operational purposes. Multirotor UAVs provide vertical lift and precise maneuverability, suitable for hovering, urban inspections, and agricultural monitoring. Fixed-wing UAVs achieve lift through aerodynamic wings, offering longer endurance and higher speed, thus suited for extended surveillance and mapping missions. Rotary-wing UAVs combine vertical lift with substantial payload capacity, ideal for logistics and search-and-rescue operations. Hybrid UAVs merge fixed-wing efficiency with vertical takeoff capabilities, enhancing operational flexibility. Lastly, flapping-wing UAVs utilize wing movements for lift and propulsion, beneficial for stealth and agile maneuvers in complex environments.
	
	\subsection{Perception Systems}
	The autonomous operation of UAVs relies on integrated perception systems, which enable effective real-time sensing and navigation in complex environments. Key sensor components typically include optical and infrared cameras, GNSS systems, LiDAR, IMUs, and millimeter-wave radars. Optical and infrared cameras provide essential visual and thermal imaging capabilities for various monitoring and inspection tasks. GNSS systems facilitate precise positioning and navigation in open environments, while LiDAR offers detailed 3D mapping and obstacle avoidance capabilities, particularly in areas without GNSS coverage. IMUs support stable flight control through attitude and motion sensing, and millimeter-wave radars enhance environmental awareness and target tracking even under adverse conditions. Together, these sensors ensure robust autonomous perception and operational reliability for UAVs.
	
	\subsection{Communication Systems}
	The communication system of UAVs plays a crucial role in information exchange. In indoor scenarios, WiFi has become the mainstream solution due to its stable transmission performance, ease of deployment, and low infrastructure costs. In urban low-altitude flight environments, although the cellular base station network can provide wide coverage, the current base station antennas are primarily designed to serve ground users, resulting in coverage blind spots and unstable signal quality when providing communication services to UAVs. In the field environments such as forests and mountainous areas, where the deployment of ground infrastructure is insufficient, the reliability of UAV communication systems is also under challenge. Ground-based radio links, while cost-effective, face significant issues of signal attenuation and blockage in complex terrains and dense vegetation. Satellite communication systems offer distinct advantages: their independence from ground infrastructure, combined with all-weather, wide-area coverage, ensures reliable support for UAVs in the wild. Specifically, low earth orbit satellite constellations can provide millisecond-level latency and high transmission bandwidth, making them ideal for real-time applications. Although geostationary satellites have high propagation delay and limited data rate, their excellent coverage and signal reliability still have irreplaceable value in specific application scenarios.
	
	\subsection{Multimodal Large Language Models}
\begin{figure}[h]
		\centering
		\includegraphics[width=1\columnwidth]{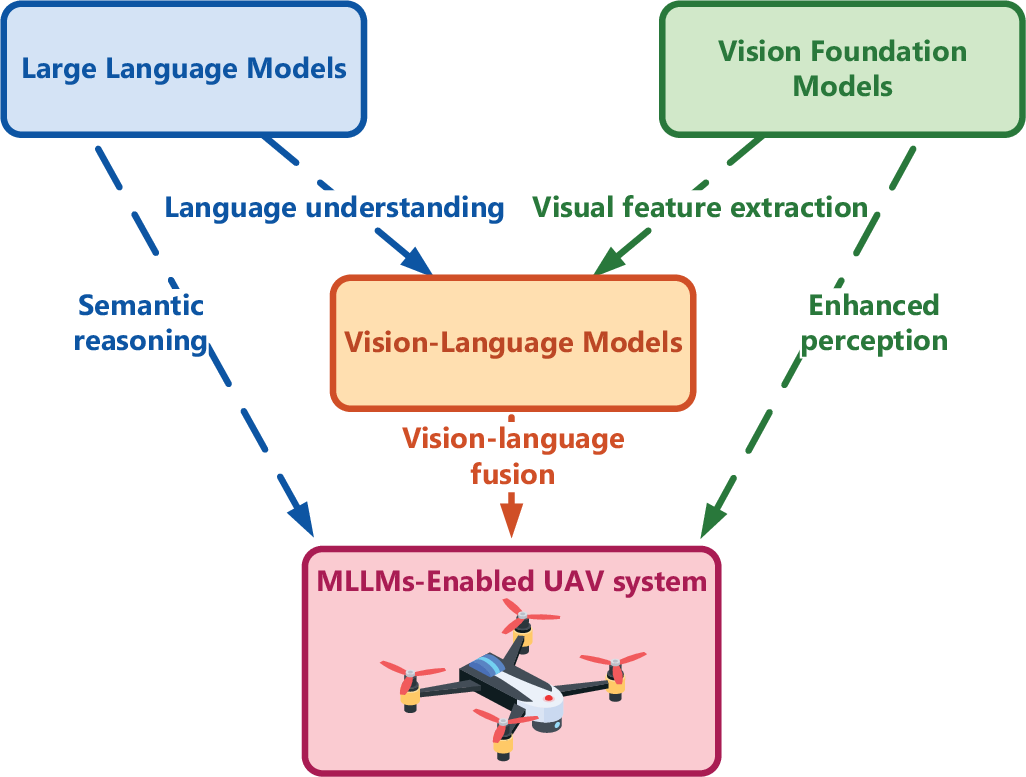}
		\caption{Synergistic contributions of LLMs, VFMs, and VLMs to an MLLMs-enabled UAV system}
		\label{mllms}
	\end{figure}

	An MLLMs-enabled UAV system consists of one or more of the following models: large language models (LLMs), vision foundation models (VFMs), and vision language models (VLMs). Specifically, LLMs specialize in processing textual data through natural language understanding and generating, making them particularly effective for semantic reasoning tasks. VFMs focus on visual data processing, demonstrating exceptional capabilities in hierarchical feature extraction from images. VLMs can process and generate visual and textual content, enabling cross-modal reasoning that enhances performance in multi-modal tasks. Fig. \ref{mllms} illustrates the synergistic contributions of LLMs, VFMs, and VLMs to an MLLM-enabled UAV system. This chapter will provide a concise overview of the current capabilities of each of these models.
	\subsubsection{Large Language Models}
	
	LLMs, such as Chat-GPT, LLaMA, and Qwen, rely on vast amounts of unlabeled text data and advanced attention mechanisms to understand context, infer meaning, and generate text. These models exhibit strong generalization abilities, particularly in zero-shot or few-shot tasks. When tackling complex problems, they decompose tasks into smaller sub-tasks and apply step-by-step reasoning, such as CoT. This ability also enables them to plan tasks and schedule actions by mobilizing the necessary resources and integrating workflows to deliver comprehensive solutions. Furthermore, these models can optimize their performance for specific tasks or domains through fine-tuning, enhancing both accuracy and efficiency in real-world applications.
	
	\subsubsection{Vision-Foundation Models}
	
	VFMs are primarily applied to computer vision, such as object detection, including object detection, semantic segmentation, and object tracking. Typical VFMs include DINO, SAM, BEiT, and MAE, which are capable of extracting diverse and highly expressive image features. Through fine-tuning or prompting, VFMs can quickly adapt to downstream tasks and have become a new paradigm in computer vision. Compared to traditional visual algorithms, VFMs typically have larger parameter scales, demonstrating outstanding generalization ability and cross-task transfer performance. Based on VFMs, systems can achieve zero-shot detection, model relationships between objects, and improve the understanding of complex scenes, significantly enhancing the performance of visual tasks.
	\begin{table*}[h]
		\centering
		\includegraphics[width=2\columnwidth]{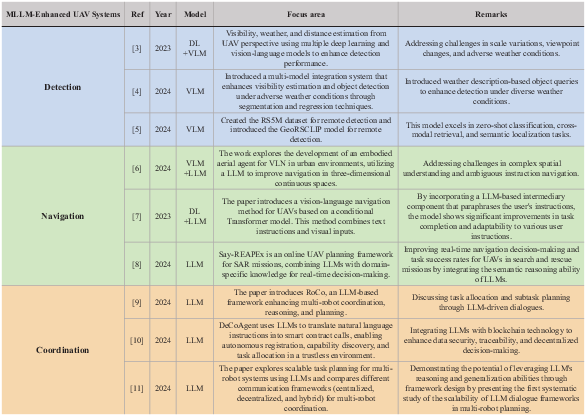}
		\caption{Related works on MLLMs-enabled UAV system.}
		\label{table}
	\end{table*}
	\subsubsection{Vision-Language Models}
	
	VLMs, such as GPT-4o, Qwen-VL, and LLaVA, are designed to tackle various tasks requiring vision and language understanding. VLMs combine the visual features extracted by VFMs with the language reasoning ability of LLMs. It is competent for multimodal tasks such as text matching, visual question answering, image description generation, etc. Compared to VFMs, VLMs further integrate language information, offering cross-modal perception abilities. Differently, VLMs extend their understanding of visual content through visual inputs in comparison with LLMs. Therefore, VLMs significantly improve performance in multi-modal tasks as an organic combination of vision and language foundation models. However, in single-modal tasks, VLMs are generally less proficient than LLMs in complex knowledge reasoning and less effective than VFMs in capturing high-resolution visual details. Furthermore, due to the complex structure and high computational cost of VLMs, efficient deployment on resource-constrained devices is also a concern.
	
	\section{Related Work}
	In this section, we review the related works of MLLMs in detection, navigation, and collaboration, and discuss the methods for the deployment of MLLMs on UAVs with limited computational resources and energy. The overview of this section is concluded in the table \ref{table}.
	
	\subsection{MLLMs-Enhanced UAV Detection System}
	Traditional UAV vision systems face challenges such as multi-scale object detection, dynamic environments, and adaptability to various application areas. Variations in flight altitude and perspective complicate detection, while the diversity of scene features limits the generalization ability of the systems. Existing methods, such as specialized training and multi-task learning, improve robustness but are costly and rely on large amounts of annotated data.
	
	MLLMs offer new solutions through zero-shot learning, contextual understanding, and multimodal integration. The author in \cite{sakaino2023dynamic} proposed an integrated system that uses various deep learning methods and VLMs to estimate visibility, distance, and weather conditions from images captured by UAV cameras. This system effectively addresses challenges such as scale variation, perspective change, and harsh weather. Reference \cite{kim2024weather} utilized VLMs to extract detailed weather, lighting, and visibility descriptions from UAV perspective images. This enhances the ability to detect from a UAV's viewpoint, making the system more resistant to adverse environmental factors such as fog, rain, and low light. Reference \cite{zhang2024rs5m} constructed a large-scale remote sensing image-text paired dataset, RS5M, and performed full fine-tuning or parameter-efficient fine-tuning of the CLIP model, developing the GeoRSCLIP model. This model excels in zero-shot classification, cross-modal retrieval, and semantic localization tasks, thereby endowing UAV detection with strong domain adaptability and task-generalization capabilities.

	\subsection{MLLMs-Enhanced Intelligent UAV Navigation Systems}
	With the introduction of MLLMs, new paradigms have emerged in UAV navigation systems. Through extensive pre-training, MLLMs can effectively learn aligned cross-modal representations, significantly enhancing task understanding and execution capabilities. Utilizing MLLMs, UAVs can integrate high-level natural language instructions and process complex real-time dynamic environmental inputs, attract tremendous interests from both academic and industrial areas.
	
	Reference \cite{zhang2024demo} proposed an urban Visual-Language Navigation (VLN) system based on UAVs, which leveraged VLMs and LLMs to enhance spatial perception and path planning capabilities, addressing challenges in complex spatial understanding and ambiguous instruction navigation. Reference \cite{chen2023vision} introduced a language-augmented cross-modal system for UAV navigation, effectively integrating multimodal features such as textual instructions and visual context. A prompt-based method was designed to address the issue of users providing diverse textual instructions within the same navigation task. This method used RESNET to extract and project environmental features and introduced an LLMs-based mediation component to reformulate user instructions, significantly improving navigation success rates. Reference \cite{doschl2024say} enhanced real-time navigation decision efficiency and task success rates in UAV search and rescue missions by integrating LLMs' semantic inference capabilities, action filtering based on the domain knowledge, and online heuristic search. It also addressed the scalability limitations of traditional LLMs based planning frameworks in dynamic environments.
	
	\subsection{Multi-Agent Coordination via MLLMs}
	In the context of multi-agent collaboration, exemplified by a UAV swarm, effectively decomposing and allocating complex tasks is key to optimize collaboration performance. Traditional methods rely on fixed task sequences or predefined task models, performing well in static or semi-static environments. However, in dynamic, rapidly changing scenarios with increasing task complexity, these methods face challenges in terms of flexibility and adaptability. Particularly, when the task environment is characterized by uncertainty or the number of agents increases, traditional methods often struggle to achieve efficient task decomposition and coordination. By leveraging the powerful dialogue generation, task inference, and knowledge integration capabilities of MLLMs, MLLMs-enabled collaborative systems not only enhance the flexibility of task allocation and planning but also improve adaptability and interpretability by integrating real-time environmental feedback.
	
	Reference \cite{mandi2024roco} explored the dialogue-driven planning capabilities of large language models in multi-robot collaborative tasks. The proposed method enabled robots to discuss task allocation and subtask planning through LLMs-driven dialogues before task execution. By integrating environmental feedback into the task planning process, LLMs further optimized task planning, enhancing adaptability and interpretability, while demonstrating scalability in human-robot collaboration scenarios. Reference \cite{jin2024decoagent} combined LLMs with blockchain technology. DeCoAgent enabled agents to autonomously register, discover each capabilities, and allocate tasks in untrusted, dynamic environments. Leveraging LLMs to transform natural language instructions into smart contract calls facilitated automated interaction between agents, AI, blockchain, and other intelligent entities, significantly improving the efficiency and adaptability of multi-agent systems in complex tasks. Reference \cite{chen2024scalable} compared decentralized multi-agent systems, centralized multi-agent systems, and hybrid multi-agent system frameworks regarding task success rates and efficiency in multi-agent collaboration. It explored how combining centralized planning and distributed feedback in a hybrid framework achieved the highest success rate and optimal scalability in various warehouse task scenarios, particularly as the number of robots increased. This work represented a systematic study of the scalability of LLMs dialogue frameworks in multi-robot planning, demonstrating the potential of leveraging LLMs' inference and generalization capabilities through framework design. It will lay a theoretical foundation for building scalable, low-cost LLMs-driven multi-robot systems in the future.
	
	\subsection{Deployment of MLLMs on UAV}	
	
	\begin{figure}[]
		\centering
		\includegraphics[width=1\columnwidth]{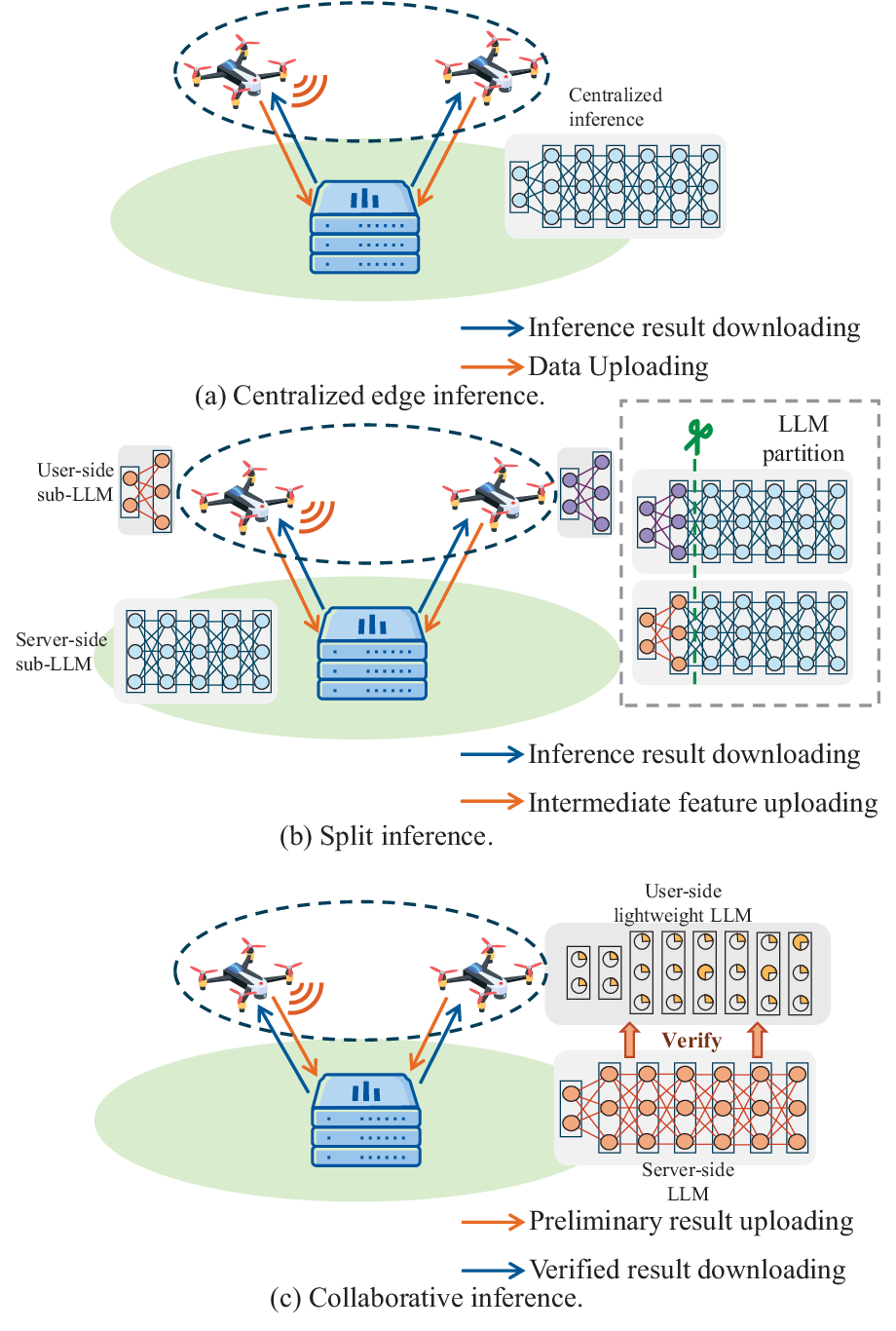}
		\caption{Three potential MLLMs inference methods when integrating MLLMs into UAVs.}
		\label{inference}
	\end{figure}
	\begin{figure*}[h]
		\centering
		\includegraphics[width=2\columnwidth]{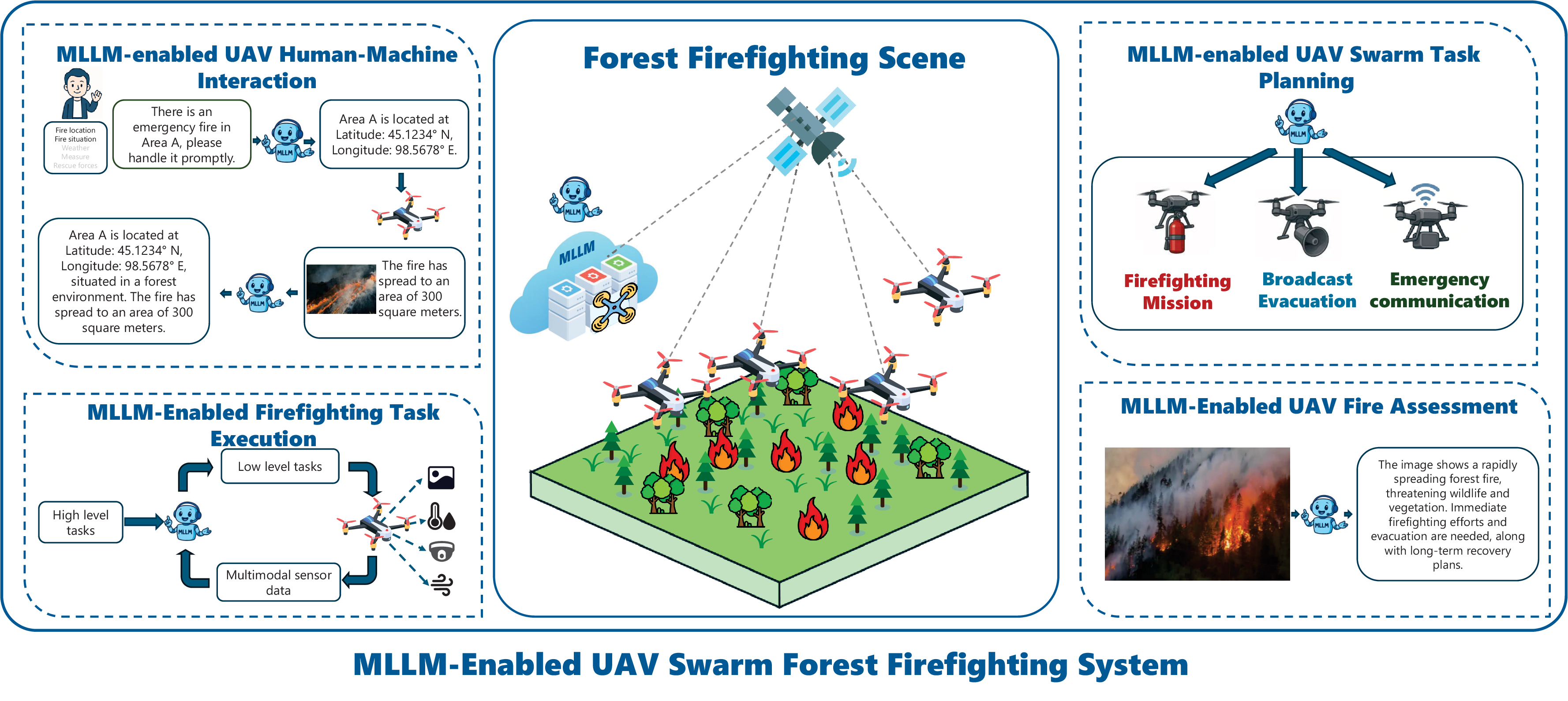}
		\caption{MLLMs-enabled UAV system in forest firefighting.}
		\label{use_case}
	\end{figure*}
	Although the deployment of MLLMs on UAVs greatly augments aerial platforms with powerful perception and inference abilities, it also poses severe constraints on on-board computation and wireless communication. Therefore, the integration of MLLMs into UAV systems has emerged as a key challenge that requires resolution. Currently, methods for deploying MLLMs on unmanned UAVs for edge inference can be classified into three categories: centralized edge inference, split inference, and collaborative inference \cite{qu2025mobile}. Centralized edge inference, which relies on wireless communication between UAVs and server-deployed MLLMs, imposes high demands on the communication quality. Reference \cite{liang2022evit} reduced the required communication overhead by decreasing the number of tokens uploaded. Other researchers considered deploying full lightweight models directly onto UAVs, but these lightweight models still faced inevitable limitations in inference capability when handling complex tasks. Additional research explored the joint deployment of MLLMs through cloud servers and UAVs. Split inference divided the complete model into two parts: one ran on the UAV, and the other ran on the server. The edge device performed initial inference on the raw data using the user-side submodel and uploaded intermediate features to the edge server for the remaining model computation \cite{shi2023task}. This split inference method was particularly suitable for MLLMs inference tasks. Compared to pure local inference, split inference offloaded most of the computation to the edge server, significantly reducing the workload on UAVs, which was critical for computation-intensive MLLMs inference tasks. Furthermore, edge network applications of MLLMs, such as mobile healthcare and autonomous driving, typically involved high sensitive private user data. Since edge devices did not share raw private data with servers, split inference mitigated privacy concerns. However, this method still required substantial communication resources due to the transmission of high-dimensional intermediate features. Reference \cite{leviathan2023fast} explored the collaborative inference method of device-server collaboration. The method of speculative decoding allowed edge devices to run smaller local MLLMs, which were known as approximate models. In comparison, the edge server deployed a larger MLLM to verify and correct the uploaded output tokens. This method effectively reduced the bandwidth requirements for inference, enhancing the possibility for UAVs to access MLLMs in infrastructure-limited environments.

	\section{Use Case: Forest Firefighting}
	In this use case, the forest fire department has deployed an MLLMs-enabled UAV swarm system to support fire detection, assessment, and extinguishment in the forest with limited infrastructures. Fig. \ref{use_case} illustrates the system architecture, where the UAV swarm gather the forest data and transmit it to the fire department management center through wireless communication links such as satellites. This section focuses on how the integration of MLLMs can address the limitations of UAV applications in firefighting by improved human-machine interaction, facilitated UAV swarm task planning, and guided fire assessment and firefighting execution.
	
	We use the open-source UAV simulation platform \textit{AirSim} to build a forest-firefighting scenario. In this scenario, all UAV scheduling and control are performed automatically by the MLLM Doubao 1.5 Vision Pro. During the mission, the MLLM assigns detection and firefighting tasks to the appropriate UAVs and generates control code from natural-language instructions. By ingesting sensor data from the UAVs, the MLLM evaluates the effectiveness of each firefighting action and dynamically updates the mission plan.  A demonstration video of the experiment is available at https://youtu.be/zwnB9ZSa5A4.
	
	\subsection{MLLMs-Enabled UAV Human-Machine Interaction}	
	
	In forest firefighting tasks, UAVs collaborate closely with human firefighters to effectively conduct firefighting and emergency response operations. To achieve this, UAVs need to comprehend human natural language commands and integrate environmental semantic information to perform intelligent reasoning and make real-time decisions, ensuring smooth and efficient cooperation with firefighters. MLLMs serve as an interface that processes natural language inputs, understanding complex instructions and allowing UAVs to respond flexibly to diverse commands. Furthermore, by incorporating RAG technology, MLLMs enable the UAVs to query a fire knowledge base for information retrieval and question answering, providing firefighters with valuable professional firefighting knowledge. Additionally, through reinforcement learning techniques, the model can proactively ask questions and address any information gaps, further enhancing the reliability and accuracy of human-machine interactions.
	
	\subsection{MLLMs-Enabled UAV Swarm Task Planning}	
	Due to the dynamic and complex nature of fire scenes, traditional UAV swarm cooperation and mission planning methods face challenges in real-time adjustments, particularly when considering factors such as the equipment carried by the UAV, remaining energy, and computational power. In contrast, MLLMs leverages the CoT framework to break down complex tasks into a series of clear, executable subtasks. This approach enables UAV swarms to achieve more efficient and precise task planning. Additionally, by utilizing the multimodal sensors of UAV swarms, MLLMs can analyze sensor data to predict fire scenarios, dynamically adjust UAV tasks, optimize resource allocation, and enhance the overall efficiency of disaster response and rescue operations.
	
	\subsection{MLLMs-Enabled UAV Fire Assessment}	
	The assessment of forest fires relies on expertise from multiple disciplines, including meteorology, ecology, and remote sensing. By fine-tuning a general pre-trained model with domain-specific knowledge for forest fire management tasks, MLLMs  can enhance its ability to integrate information and perform interdisciplinary reasoning. With this capability, UAV can efficiently fuse multi-source heterogeneous data, such as video, infrared, and meteorological information, to accurately identify the specific location, severity, and spread of fires. This enables real-time monitoring, situational awareness, and risk assessment of fire incidents.
	
	\subsection{MLLMs-Enabled Firefighting Task Execution}		
	In forest firefighting scenarios, UAVs must exhibit advanced perception and autonomous decision-making capabilities to effectively manage dynamic environmental conditions. MLLMs significantly enhance UAVs ability to accurately interpret environmental data and swiftly adapt to evolving fire situations. Unlike traditional rigid task-flow execution models, MLLMs leverage the CoT reasoning framework to dynamically reassess and replan UAV actions in response to real-time fire developments. This adaptive capability enables precise execution of critical tasks, such as deploying fire suppression agents precisely according to fire scale and intensity, or adjusting evacuation paths based on predictive analysis of fire spread directions. Consequently, MLLMs empower UAVs to perform firefighting missions with increased accuracy, efficiency, and adaptability, effectively responding to the complexities inherent in forest fire emergencies.

	\section{Challenges, Future Research Directions and Recommendations}
	Despite the promising potential of integrating MLLMs into UAV swarm systems, the research in this field is still in its early stages and facing several challenges. The following section will discuss these challenges and explore future research directions.
	\subsection{Addressing the Hallucination Issue in MLLMs for UAV system}	
	
	In MLLMs, hallucination refers to scenarios where the text or reasoning generated by the model does not align with input sensor data or conflicts with objective reality. For UAV systems, such hallucinations can result in crashes or collateral damage, which leaves far less margin for error than chat-based applications. Moreover, hallucinations are even more likely because of cross-modal misalignment caused by factors such as the downward perspective of cameras, varying weather and lighting conditions, and the scarcity of training data. To address this, task-specific datasets should be incorporated into both the data pipeline and the model fine-tuning stage, ensuring that MLLMs exhibit low hallucination rates for common tasks before deployment. At the inference level, it is necessary to prompt the model to output confidence scores for its reasoning, employing multi-run agreement checks or fallback safety strategies for low-confidence decisions. At the system level, using high-precision redundant sensors and gathering multi-angle, multi-temporal observations can partly offset hallucinations induced by environmental noise. These enhancements, however, increase both cost and mission duration and therefore force a trade-off between operational efficiency and overall reliability.
	
	\subsection{Balancing Inference Latency and CoT Length}	
	
	Balancing inference latency and CoT length is a critical challenge in the MLLMs-enabled UAV swarm.  A longer CoT can improve reasoning accuracy, mitigate hallucination, and enhance problem-solving in complex scenarios. However, an excessively long CoT inevitably increases the inference latency, which is unacceptable for time-critical swarm missions with tight energy budgets. To address this challenge, future research should adaptively adjust the CoT length based on a comprehensive evaluation of task complexity, real-time constraints, and available computing resources. At the same time, suitable early-exit mechanisms during inference can be introduced to reduce the generation of superfluous tokens. A complementary approach is to adopt a hierarchical inference pipeline in which the on-UAV model uses a short CoT to guarantee real-time responses, while the full CoT is processed on ground equipment to verify those results, striking an optimal balance between latency and accuracy.
	
	\subsection{Optimizing Communication and Computation Efficiency}	
	Optimizing communication and computation efficiency remains a significant challenge in MLLMs-enabled UAV swarm systems. UAVs inevitably have to carry out outdoor missions, yet low-altitude base station coverage outdoors often contains blind spots, making it difficult to provide reliable, high-bandwidth communication links. This limitation makes it hard to meet the stringent requirements of centralized MLLM inference commonly used by indoor robots. Therefore, from the communications perspective, future research could explore collaboration between low earth orbit satellites, geostationary satellites, and base stations to provide UAVs with stable and wide-area coverage. Additionally, research could investigate semantic communication, where transmitting highly compressed semantic features instead of raw data can significantly reduce communication overhead. From the computation perspective, UAV platforms face constraints in computational resources and energy.  Exploring lightweight MLLM models, such as through knowledge distillation, quantization, and pruning, can make inference better suited for resource-constrained UAV computing platforms. Progressive or early-exit inference methods could also be implemented, enabling flexible adjustment of the inference process based on real-time communication conditions.

	\section{Conclusion}
	
	MLLMs-enabled UAV swarm can revolutionize the current paradigm of UAV mission execution. With the rapid development of the low-altitude economy, MLLMs will empower UAVs with breakthrough capabilities in target detection, autonomous perception and navigation, task inference and allocation, and multi-agent collaboration, thereby promoting the application of UAVs in a wider range of scenarios. This paper introduces the fundamental architectures and functions of UAVs and MLLMs, analyzes how MLLMs can enhance UAV system performance in terms of target detection, autonomous navigation, and multi-agent coordination, and explores methods for integrating MLLMs into UAV systems. To demonstrate the practicality and effectiveness of this integrated approach, we present a use case focusing on forest firefighting operations. Finally, we discuss the challenges, outline future research directions, and provide strategic recommendations to further advance the development and deployment of MLLMs-enabled UAV systems.
	
	\bibliographystyle{IEEEtran}
	\bibliography{reference}

\end{document}